\documentclass{article}

\usepackage{PRIMEarxiv}

\usepackage[utf8]{inputenc} 
\usepackage[T1]{fontenc}    
\usepackage{float}
\usepackage{hyperref}       
\usepackage{url}            
\usepackage{booktabs}       
\usepackage{amsfonts}       
\usepackage{nicefrac}       
\usepackage{microtype}      
\usepackage{lipsum}
\usepackage{fancyhdr}       
\usepackage{graphicx} 
\usepackage{tikz}
\usetikzlibrary{positioning, arrows.meta}
\usepackage{amsmath} 
\graphicspath{{media/}}     
\usepackage{graphicx}
\usepackage{multirow}
\usepackage{amsmath,amssymb,amsfonts}
\usepackage{amsthm}
\usepackage{mathrsfs}
\usepackage[title]{appendix}
\usepackage{xcolor}
\usepackage{textcomp}
\usepackage{manyfoot}
\usepackage{booktabs}
\usepackage{algorithm}
\usepackage{algorithmicx}
\usepackage{algpseudocode}
\usepackage{listings}
\usepackage{tikz}
\usetikzlibrary{positioning, arrows.meta}

\usepackage{float}
\usepackage{longtable}
\usepackage{threeparttable}
\usepackage{hyperref}
\usepackage{url}
\usepackage[utf8]{inputenc}
\usepackage[T1]{fontenc}

\usepackage{cleveref}

\theoremstyle{thmstyleone}

\theoremstyle{thmstyletwo}

\theoremstyle{thmstylethree}

\title{TT-DAC-PS: Twin-Target Deterministic Actor-Critic with Policy Smoothing for Optimal Trade Execution
}

\author{
  Ilia Zaznov, Atta Badii \\
  Department of Computer Science \\
  University of Reading \\
  Reading, UK \\
  \texttt{\{i.zaznov@pgr.reading.ac.uk, atta.badii@reading.ac.uk\}} \\
  \And
  Julian Kunkel \\
  Department of Computer Science/GWDG \\
  University of Göttingen \\
  Goettingen, Germany \\
  \texttt{julian.kunkel@gwdg.de} \\
  \And
  Alfonso Dufour \\
  ICMA Centre, Henley Business School \\
  University of Reading \\
  Reading, UK \\
  \texttt{a.dufour@icmacentre.ac.uk} \\
}

\begin{document}
\maketitle

\begin{abstract}
This study addresses the optimal execution of large stock sell programs by introducing TT-DAC-PS (Twin-Target Deterministic Actor-Critic with Policy Smoothing), a deterministic actor-critic architecture that combines: (i) twin exponential-moving-average critic targets with pessimistic min backup, (ii) TD3-style target policy smoothing noise, (iii) delayed actor updates, and (iv) conservative Q regularisation to curb overestimation. Exploration uses Ornstein–Uhlenbeck (OU) noise with a hybrid schedule: deterministic episode-wise decay, variance-guided adjustment based on recent reward dispersion, and a Soft Actor–Critic (SAC)-style temperature that is learned and mapped to the noise scale. The environment integrates Almgren--Chriss (AC) trade impact with Limit Order Book (LOB) prices and volumes, normalised state features, per-step volume participation caps, and a utility-based reward. Our new trade execution algorithm is applied to LOB data for ten U.S. stocks. The performance is assessed with respect to reinforcement-learning baseline algorithms such as the Proximal Policy Optimisation (PPO), Soft Actor–Critic (SAC), and Advantage Actor–Critic (A2C), as well as alternative trade execution algorithms such as Time-weighted average price (TWAP), volume-weighted average price (VWAP), and AC. The proposed model consistently reduces mean implementation shortfall percentage with competitive variance, outperforming classical baselines and standard reinforcement learning benchmark models. Ablation experiments further indicate that both the Twin-Target pessimistic backup and the hybrid adaptive exploration design contribute to the observed improvements, with the largest degradation appearing in more challenging instruments when these stabilisation mechanisms are removed. Results indicate that adaptive exploration is effective in episodic execution with time-varying liquidity and non-stationary spreads, where fixed-noise schedules are brittle. 

\end{abstract}

\keywords{
Optimal Trade Execution \and
Implementation Shortfall \and
Reinforcement Learning \and
Deterministic Actor--Critic \and
TD3-Style Policy Smoothing \and
Adaptive Exploration \and
Ornstein--Uhlenbeck Noise \and
Almgren--Chriss Model \and
Limit Order Book (LOB) \and 
}

\section{Introduction}
At its core, optimal execution is about turning a trading intention into realised trades with minimal cost and risk. When a portfolio manager decides to sell a large number of shares, the goal is to complete that sale without unduly moving the market price or incurring excessive transaction costs, while also controlling the uncertainty of outcomes. A central measure of performance is the implementation shortfall (IS): the difference between the paper value of the order at the time the decision is made and the actual amount realised once the order is executed. Intuitively, IS captures the total slippage of execution relative to the decision price, including both explicit costs (e.g., fees, bid–ask spread) and implicit costs (e.g., market impact and timing). In percentage terms, a lower IS means the realised proceeds are closer to the notional value at the decision time, signalling more efficient execution.

From this simple objective—selling a block of shares while keeping IS low—several practical challenges emerge. First, large orders can move the market price, so trading too aggressively may depress execution prices (market impact). Second, even though a large order can be broken down into smaller trades executed over time, trading too slowly exposes the order to adverse price movements (timing risk). Third, market conditions evolve intraday: liquidity waxes and wanes, spreads widen or tighten, and volatility changes with macro news or microstructure states. These realities create a tension between speed and cost and motivate schedules that adapt to market conditions.

Classical solutions address this trade-off with simple, robust trading schedules. The Time-Weighted Average Price (TWAP) execution strategy spreads the order evenly across the trading horizon, while the Volume-Weighted Average Price (VWAP) algorithm aligns trading with expected or observed volume. The Almgren–Chriss (AC) framework formalises the mean–variance trade-off between expected shortfall and execution risk, modelling both temporary and permanent price impacts to derive an optimal trading trajectory under specified assumptions. These approaches remain widely used because they are interpretable, controllable, and resilient to mild parameter uncertainty \cite{almgren2001optimal, almgren2003optimal, gatheral2011optimal}. However, the AC approach is fragile when facing structural misspecification, such as non-linear impact \cite{wang2015optimal}, time‑varying liquidity \cite{cartea2015algorithmic}, structural breaks and regime shifts, or non-stationary bid-ask spreads \cite{mastromatteo2014agent}.

Real markets are non-stationary. Several papers show that adaptive algorithms outperform AC in non-stationary markets \cite{nevmyvaka2006reinforcement, hendricks2014reinforcement, fang2021universal}. Liquidity and volatility are time-varying, spreads respond to order flow and news, and impact can be nonlinear or state-dependent. Fixed schedules or static parameterisations can therefore underperform when conditions shift within an episode. In this setting, adaptive methods that learn to respond to changing microstructure conditions are appealing—provided they remain stable, respect safety constraints, and deliver consistent outcomes across instruments and regimes \cite{cartea2021deep, hafsi2024optimal, macri2024reinforcement, tonkin2025benchmarking}. Beyond linear-impact formulations, richer models of limit order book dynamics further highlight the importance of liquidity structure in execution design. For instance, \cite{alfonsi2009optimal} shows that optimal strategies depend on the shape of the order book, reinforcing the limitations of static or overly simplified impact assumptions. Reinforcement learning has emerged as a natural framework for such adaptive control problems, enabling policies that respond dynamically to evolving market conditions while handling high-dimensional state representations \cite{hambly2021recent}. Recent advances extend this paradigm to continuous-action and online execution settings, where policies must operate under strict constraints and stochastic environments \cite{micheli2024deep}.

This paper proposes a deep reinforcement learning approach tailored to these requirements. The TT-DAC-PS model is presented that learns a continuous action policy for share allocation at each step, operating within an execution environment that blends AC-style impact with Limit Order Book (LOB)–derived signals. 

Exploration in continuous-control reinforcement learning is highly sensitive to the choice and scheduling of noise, which can lead to unstable learning and inconsistent performance across environments, particularly in financial applications with non-stationary dynamics \cite{lillicrap2015continuous, hambly2021recent}. To address this issue, the proposed TT-DAC-PS framework employs an adaptive OU noise process, where the exploration scale is dynamically adjusted based on training progress and reward variability, improving stability and consistency across different trading activity regimes. OU noise is modulated by (a) a deterministic decay, (b) a variance-aware controller, and (c) a learned SAC-style temperature mapped to the noise scale, all under strict participation caps and a terminal liquidation rule. This combination preserves stability while enabling the model to probe new behaviours when learning appears to plateau. The noise scale is bounded within \([\sigma_{\min}, \sigma_0]\) to maintain safety.

The environment provides normalised states (including time-to-go, remaining inventory, recent returns, and volume), AC-parameterised execution pricing with spread and impact, and explicit safety measures via per-step participation caps. Reward shaping leverages AC utility to encourage trajectories that reduce both expected shortfall and risk in a stepwise manner, keeping learning focused on improvements in execution quality. Robust LOB preprocessing—Median Absolute Deviation (MAD) filtering for prices and interquartile range (IQR) capping for volumes—stabilises inputs that parameterise cost and risk.

The proposed model is evaluated on ten U.S. stocks (ADBE, AMD, AVGO, BKR, CSCO, CMCSA, ORCL, PEP, PYPL, INTC) and compared against strong deep RL baselines (PPO, SAC, A2C) and classical schedules (TWAP, VWAP, AC). Across instruments, the model achieves a consistently low mean IS(\%) with competitive variance, improving on classical baselines and deep RL baselines.

\section{Related Work}

\subsection{Overview and Scope}
\begin{figure}[bt]
\centering
\begin{tikzpicture}[
  grow=right,
  edge from parent/.style={draw,->,thick},
  parent anchor=east,
  child anchor=west,
  box/.style={
    font=\scriptsize,
    rectangle,
    draw,
    align=center,
    text width=1.5cm,      
    minimum height=0.9cm, 
    inner sep=2pt
  },
  root/.style={box, fill=gray!15, font=\normalsize},
  top/.style ={box, fill=#1!10},
  leaf/.style={box, fill=#1!5},
  level 1/.style={level distance=3.2cm, sibling distance=3cm},
  level 2/.style={level distance=2cm, sibling distance=1cm},
  level 3/.style={level distance=3.2cm, sibling distance=1.5cm}
]

\node[root] (root) {Optimal Execution Research}
  child { node[top=blue] {Classical Execution Models}
    child { node[leaf=blue] {TWAP / VWAP} }
    child { node[leaf=blue] {Almgren--Chriss} }
    child { node[leaf=blue] {Robust / Portfolio} }
  }
  child { node[top=green] {Market Microstructure \& LOB}
    child { node[leaf=green] {Order Flow, Spread, Depth} }
    child { node[leaf=green] {Statistical Features} }
    child { node[leaf=green] {Impact Decay Models} }
  }
  child { node[top=red] {Reinforcement Learning}
    child { node[top=yellow] {Exploration Strategies}
      child { node[leaf=yellow] {OU Noise} }
      child { node[leaf=yellow] {Parameter / Network Noise} }
      child { node[leaf=yellow] {Variance-Aware / Ensemble} }
    }
  }
  child { node[top=purple] {Impact Modeling \& Simulators}
    child { node[leaf=purple] {AC-style Simulators} }
    child { node[leaf=purple] {LOB-based Simulators} }
    child { node[leaf=purple] {Cross-Impact Models} }
  }
  child { node[top=orange] {Evaluation \& Risk}
    child { node[leaf=orange] {Implementation Shortfall} }
    child { node[leaf=orange] {CVaR / Tail Risk} }
    child { node[leaf=orange] {Performance Attribution} }
  };

\end{tikzpicture}
\caption{Taxonomy of optimal execution research.}
\label{fig:taxonomy}
\end{figure}
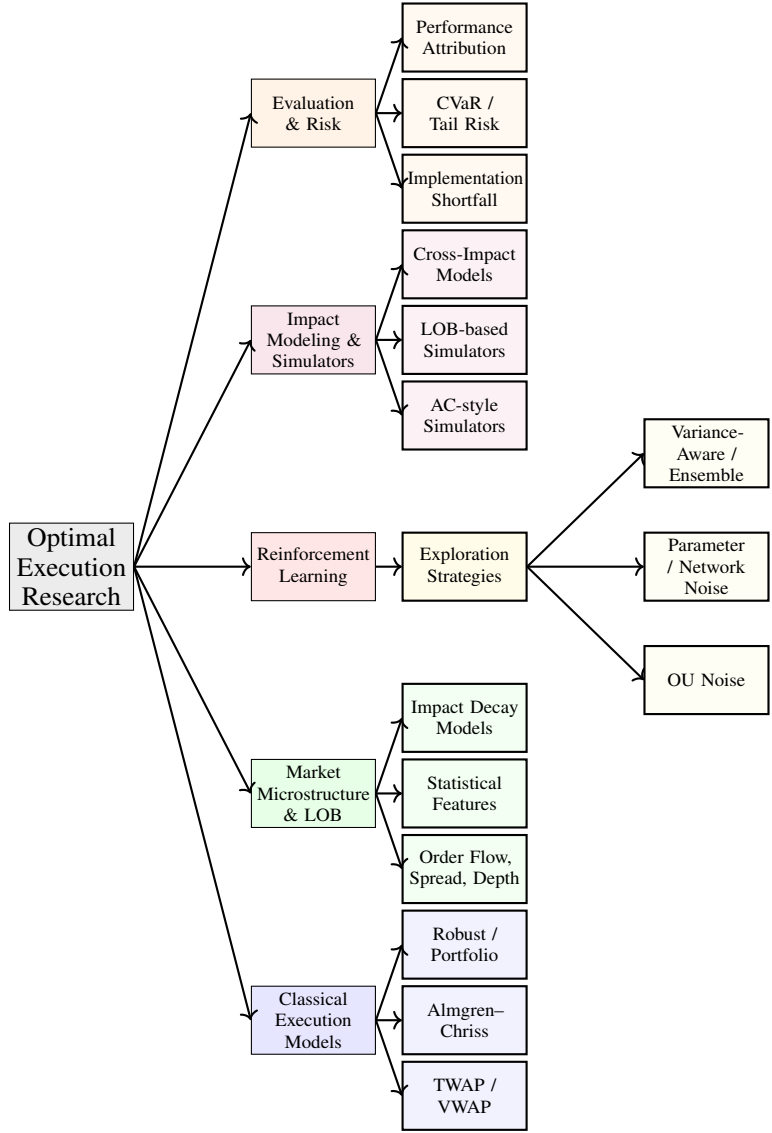

This section surveys the strands that materially shape execution design and evaluation, using the taxonomy in Figure~\ref{fig:taxonomy} as an organising lens rather than an end in itself. The focus is on: (i) classical execution theory, which defines the cost–risk frontier and the structural role of permanent vs. temporary impact; (ii) market microstructure and LOB evidence that constrains modelling choices (concave impact, impact decay, clustered order flow) and motivates state variables; (iii) reinforcement learning methods for adaptive scheduling under safety constraints, with emphasis on exploration design and stability; (iv) impact models and simulators that trade off interpretability and realism, determining what performance “means” in-sample; and (v) evaluation and risk frameworks that align research with deployment (IS, tail risk, attribution).

Our objective is not to catalogue algorithms but to connect assumptions to consequences: how a choice about impact (linear vs. transient), state (LOB features vs stylised proxies), or exploration (fixed vs. variance-aware) propagates to execution quality, robustness, and comparability across instruments. The taxonomy highlights these dependencies: classical models set the normative baseline; microstructure evidence narrows plausible dynamics; RL proposes adaptive mechanisms; simulators instantiate testable worlds; and evaluation criteria adjudicate trade-offs. This structure was adopted to show what carries over in practice and where claims depend on simulator design or fragile exploration choices.
 
\subsection{Classical Optimal Execution}
The classical literature on optimal execution provides the mathematical backbone for trade scheduling, focusing on the interplay between market impact, price risk, and trading speed. The Almgren--Chriss (AC) framework \cite{almgren2001optimal,bertsimas1998optimal} is seminal, modelling the expected cost and risk of liquidating a position over a discrete horizon. The AC model assumes price dynamics follow a Brownian motion and that trades impact prices both temporarily and permanently:

\[
p_{t+1} = p_t - \gamma x_t + \sigma \sqrt{\tau} \epsilon_t
\]
\[
p_t^{\text{exec}} = p_t - \epsilon \cdot \operatorname{sign}(x_t) - \frac{\eta}{\tau}x_t
\]

where \(x_t\) is the trade at time \(t\), \(\gamma\) is permanent impact, \(\eta\) is temporary impact, \(\epsilon\) is the half-spread, and \(\tau\) is the time interval. The trader seeks to minimise a mean-variance objective:

\[
\min_{x_{0:N-1}} \quad \mathbb{E}[\text{Cost}] + \lambda \cdot \mathrm{Var}[\text{Cost}],
\]
where the implementation shortfall cost is defined as
\[
\text{Cost} = Q p_0 - \sum_{t=0}^{N-1} x_t \left( p_t - \epsilon \cdot \operatorname{sign}(x_t) - \frac{\eta}{\tau}x_t \right),
\]
so that
\[
\mathbb{E}[\text{Cost}] = \mathbb{E}\!\left[ Q p_0 - \sum_{t=0}^{N-1} x_t \left( p_t - \epsilon \cdot \operatorname{sign}(x_t) - \frac{\eta}{\tau}x_t \right) \right],
\]
and
\[
\mathrm{Var}[\text{Cost}] = \sigma^2 \sum_{t=0}^{N-1} x_t^2.
\]

The solution yields a trade trajectory that balances front-loading (to reduce risk) and smoothing (to reduce impact), controlled by the risk aversion parameter \(\lambda\).

TWAP and VWAP are practical, robust baselines:
\begin{itemize}
\item TWAP (Time-Weighted Average Price) distributes trades evenly:
  \[
  x_t^{\text{TWAP}} = \frac{Q}{N}
  \]

where \(x_t\) is the trade size at time \(t\), and \(Q\) denotes the total initial inventory to be executed over the trading horizon, i.e., \(Q = \sum_{t=0}^{N-1} x_t\).
\item VWAP (Volume-Weighted Average Price) aligns trades with volume:
  \[
  x_t^{\text{VWAP}} = Q \cdot \frac{v_t}{\sum_{k=0}^{N-1} v_k}
  \]
where \(v_t\) is expected or observed volume.
\end{itemize}

No-dynamic-arbitrage constraints \cite{gatheral2010no} require the permanent impact to be linear in aggregate order flow, ruling out price manipulation and ensuring market integrity. Extensions to the AC model include transient impact and resilience, where impact decays as liquidity replenishes \cite{gatheral2013transient}. The propagator model expresses the transient impact as:

\[
p_t = p_0 + \sum_{k=1}^t G(k) x_{t-k}
\]
where \(G(k)\) is a decay kernel.

Robust execution hedges against uncertainty in volatility and impact, often via minimax or distributionally robust formulations \cite{schied2013robust}. Portfolio-based models connect optimal execution to dynamic trading and signal decay \cite{garleanu2013dynamic}, embedding execution in broader investment processes.

\subsection{Market Microstructure and the Limit Order Book}
The Limit Order Book (LOB) is the microstructural engine of modern electronic markets, mediating liquidity, price formation, and execution outcomes. LOB models capture queueing dynamics, order types, and the evolution of spreads and depth \cite{gould2013limit}. Statistical analyses reveal heavy-tailed distributions of order sizes, persistent autocorrelations in order flow, and clustering of volatility \cite{bouchaud2002statistical,cont2011statistical}. Empirical response-function studies further show that price changes reflect a complex mixture of exogenous information and endogenous order-flow dynamics \cite{bouchaud2004fluctuations}.

The LOB state at time \(t\) can be described by:

\[
\text{LOB}_t = \{(p_i^{\text{bid}}, v_i^{\text{bid}}), (p_j^{\text{ask}}, v_j^{\text{ask}})\}_{i,j}
\]

Key LOB features include:
\begin{itemize}
\item \textbf{Imbalance:}
\[
  \text{Imbalance}_t = \frac{V^{\text{bid}}_t - V^{\text{ask}}_t}{V^{\text{bid}}_t + V^{\text{ask}}_t}
\]
where \(V^{\text{bid}}_t\) and \(V^{\text{ask}}_t\) denote the available volumes at the best bid and ask quotes, respectively, at time \(t\).

\item \textbf{Spread:}
\[
s_t = p^{\text{ask}}_{1,t} - p^{\text{bid}}_{1,t},
\]
where \(p^{\text{ask}}_{1,t}\) and \(p^{\text{bid}}_{1,t}\) are the best ask and best bid prices (level 1) at time \(t\).

\item \textbf{Depth:} Sum of volume at the best bid and ask quotes:
\[
\text{Depth}_t = V^{\text{bid}}_t + V^{\text{ask}}_t.
\]
\end{itemize}

Self-exciting point processes (Hawkes processes) model clustered arrivals and the endogenous nature of volatility \cite{bacry2015hawkes,hardiman2013critical}. Empirical studies show that market impact is often concave (e.g., square-root law) and decays over time \cite{toth2011anomalous,bershova2013non,almgren2005direct}. Transient impact models reconcile these facts:

\[
I_t = \sum_{k=0}^{K} G(k) x_{t-k}
\]

Microstructure-driven strategies adapt execution to current LOB conditions, using real-time imbalance, spread, and volatility signals to modulate participation rates.

\subsection{Reinforcement Learning for Trading and Execution}
Reinforcement Learning (RL) offers a data-driven approach to adaptive execution, learning policies that respond to market states and constraints. Early RL applications in execution used tabular methods to adapt schedules to liquidity proxies \cite{nevmyvaka2006reinforcement}. Modern deep RL methods such as DDPG \cite{lillicrap2015continuous}, TD3 \cite{fujimoto2018addressing}, SAC \cite{haarnoja2018soft}, and PPO \cite{schulman2017proximal} enable continuous control and scalable learning.

The RL agent interacts with the environment, observing state \(s_t\), taking action \(a_t\), and receiving reward \(r_t\). The objective is to maximise expected cumulative reward:

\[
J(\theta) = \mathbb{E}_{s_0, a_0, \dots} \left[ \sum_{t=0}^{N-1} r_t \right]
\]

In actor-critic methods, the actor is a deterministic policy \(\mu_\theta(s)\), parameterised by \(\theta\), which maps states to actions, i.e., \(a = \mu_\theta(s)\). The critic estimates the action-value function \(Q(s,a)\), defined as the expected cumulative reward starting from state \(s\), taking action \(a\), and thereafter following the policy \(\mu_\theta\). Policy updates use deterministic policy gradients \cite{silver2014deterministic}:

\[
\nabla_\theta J(\theta) \approx \mathbb{E}_{s} \left[ \nabla_a Q(s,a) |_{a = \mu_\theta(s)} \nabla_\theta \mu_\theta(s) \right]
\]

Reward shaping is critical for stable learning. In execution, reward typically aligns with a reduction in implementation shortfall or AC utility:

\[
r_t = \frac{U(q_{t-1}) - U(q_t)}{U(q_{t-1})}
\]
where \(U(q) = E(q) + \lambda V(q)\).
\(U(q)\) denotes the Almgren--Chriss utility at remaining inventory \(q\), defined as the sum of the expected execution cost \(E(q)\) and the risk term \(V(q)\) weighted by the risk-aversion parameter \(\lambda\).

Recent work demonstrates that RL can outperform static schedules, especially in non-stationary environments, but requires careful handling of constraints and exploration \cite{cartea2021deep, hafsi2024optimal, macri2024reinforcement,ning2018double,lin2021end,yang2020deep,lin2020deep}. Comprehensive surveys further highlight the rapid progress and practical relevance of RL methods in financial decision-making under uncertainty \cite{hambly2021recent}.

\subsection{Exploration Strategies in Continuous Control}
Exploration is a key challenge in RL for execution, where unsafe actions can violate participation caps or trigger excessive impact. DDPG employs Ornstein--Uhlenbeck (OU) noise for temporally correlated exploration \cite{lillicrap2015continuous}:

\[
\xi_{t+1} = \theta (\mu - \xi_t) dt + \sigma dW_t
\]
where \(\xi_t\) is noise, \(\theta\) is mean reversion, \(\mu\) is mean, \(\sigma\) is volatility, and \(dW_t\) is Brownian motion.

Alternative approaches inject noise in parameter space \cite{plappert2017parameter}:

\[
\theta' = \theta + \mathcal{N}(0, \sigma^2 I)
\]

or use noisy neural networks \cite{fortunato2017noisy}, \cite{burda2018exploration}, that inject stochasticity directly into the policy. Intrinsic motivation and curiosity bonuses encourage exploration in sparse-reward settings \cite{pathak2017curiosity}.

Ensemble-based methods, including bootstrapped Q-learning, sample implicit a-posteriori states for deep exploration \cite{osband2016deep}.

Another line of work draws on multi-armed bandit theory, where exploration is modulated based on uncertainty or reward variability \cite{russo2018tutorial}. In deep reinforcement learning, entropy regularisation serves a similar role by stabilising policy updates and maintaining sufficient exploration, particularly in non-stationary environments \cite{haarnoja2018soft, schulman2017proximal}.

In the context of optimal execution, exploration must be carefully controlled. Excessive exploration may lead to unstable trading behaviour or overfitting to transient market conditions. To address this, adaptive exploration schedules that respond to recent reward variability can help maintain a balance between exploration and stability.

\subsection{Impact Modelling and Simulators}
Simulators are essential for benchmarking execution strategies. AC-style simulators offer interpretable, closed-form impact models \cite{almgren2001optimal,cartea2021deep}, supporting analytic baselines and controlled experiments. To increase realism, transient impact kernels and resilience mechanisms are integrated \cite{gatheral2010no,donier2015makes}, reproducing empirically observed impact decay and recovery:

\[
p_t = p_0 + \sum_{k=1}^t G(k) x_{t-k}
\]

Cross-impact models extend single-asset frameworks to multi-asset settings, capturing spillover effects \cite{schneider2019cross,benzaquen2017crossimpact,mastromatteo2017tradinglightly}.

High-fidelity multi-agent simulators such as ABIDES provide an event-driven market microstructure testbed for evaluating execution and learning-based agents under realistic matching, latency, and interaction effects \cite{byrd2019abides,byrd2020abides}.

Data-driven LOB simulators use point processes or neural surrogates to model high-frequency dynamics and conditional impact \cite{zaznov2022predicting, zaznov2024intraday, sirignano2019deep,zhang2019deeplob}. 

Recent deep architectures incorporate attention mechanisms to improve LOB representation learning and forecasting performance, further motivating feature design in LOB-driven decision problems \cite{jung2024attentionlob}.

Recent testbeds blend stylised impact with empirical LOB features, enabling fair comparisons across methods and ablations over liquidity regimes, latency, and constraint handling \cite{schnaubelt2020optimal,cartea2021latency}.

\subsection{Evaluation and Risk}
Implementation shortfall (IS) is the canonical metric for execution quality, capturing both explicit (fees, spread) and implicit (impact, timing) costs \cite{perold1988implementation}:

\[
\mathrm{IS}(\%) = 100 \cdot \frac{Q p_0 - \sum_t x_t p_t^{\text{exec}}}{Q p_0}
\]

where \(Q p_0\) is the notional portfolio value at decision time, given by the initial inventory \(Q\) times the initial mid-price \(p_0\).

Risk is often quantified via variance or tail metrics such as Conditional Value-at-Risk (CVaR) \cite{rockafellar2000optimization}:

\[
\mathrm{CVaR}_\alpha(X) = \mathbb{E}[X \mid X \geq \mathrm{VaR}_\alpha(X)]
\]

Performance attribution considers signal decay, inventory trajectories, and state-dependent liquidity \cite{garleanu2013dynamic}. Production readiness demands stress tests over regime shifts, volatility spikes, and liquidity droughts, echoing microstructure evidence on clustered order flow and time-varying depth \cite{cont2011statistical,cartea2015algorithmic}.

\subsection{Positioning}
Compared to prior RL execution work \cite{nevmyvaka2006reinforcement,cartea2021deep,ning2018double,osband2016deep}, this paper targets exploration as a first-class design variable for episodic, non-stationary execution. A simple, model-agnostic hybrid schedule is proposed: OU-correlated action noise with amplitude modulated by recent reward dispersion, bounded by strict participation and inventory caps. This mechanism integrates seamlessly with actor--critic methods such as DDPG and SAC \cite{lillicrap2015continuous,haarnoja2018soft}, draws on variance-aware insights from bandits and ensemble exploration \cite{russo2018tutorial,osband2016deep,plappert2017parameter,burda2018exploration}, and preserves safety via hard constraints. 

For evaluation, simulators are employed that blend AC-style interpretability with transient impact and empirically motivated LOB features \cite{almgren2001optimal,gatheral2013transient,cartea2015algorithmic,schnaubelt2020optimal,cartea2021latency}, and both mean shortfall and its standard deviation (as a variability measure) are reported to align with deployment standards \cite{perold1988implementation,rockafellar2000optimization}. This positioning provides a bridge between the rigour of classical execution and the adaptivity of deep RL in environments characterised by non-stationary liquidity and clustered volatility.

\section{Methodology}\label{sec:method}

\subsection{Overview}\label{subsec:overview}
Our methodology integrates deterministic policy gradient learning with a simulator that blends the tractability of the Almgren--Chriss (AC) cost-risk formulation and the empirical realism of limit order book (LOB) features. The central objective is episodic liquidation of a fixed inventory within a finite horizon under the dual pressures of market impact and timing risk while respecting operational safety constraints. A state is designed as a representation that captures the determinants of execution quality—time-to-go, inventory, recent price dynamics, and trading activity—then learns a continuous policy that maps these states to safe sell allocations capped by per-step participation limits. Exploration is treated as a first-class design component: rather than relying on a brittle, fixed noise schedule, a hybrid adaptive scheme is employed that modulates Ornstein–Uhlenbeck (OU) action noise through a deterministic episode-wise decay and a variance-aware adjustment driven by recent episode returns. This enables the policy to probe new behaviours when learning stagnates while reducing noise in volatile phases, without ever violating participation caps or inventory feasibility. The reward function uses AC utility at each step, ensuring that learning signals are tightly coupled to the execution objective and not only to a terminal shortfall. Training leverages experience replay, soft target networks, and robust preprocessing of LOB inputs via Median Absolute Deviation (MAD) filtering for prices and Inter Quartile range (IQR) capping for volumes. Evaluation is conducted with deterministic policies and identical constraints across all methods to ensure comparability, reporting the mean implementation shortfall in percentage terms and dispersion across episodes.

\subsection{Key Contributions}\label{subsec:contributions}
Our methodology builds on established reinforcement learning approaches to optimal execution, combining deterministic policy gradient methods with a simulator grounded in the Almgren--Chriss (AC) cost-risk framework and enriched with limit order book (LOB) features. Prior work has shown that actor--critic methods can effectively learn execution policies under market impact and stochastic price dynamics, while AC-based formulations provide a tractable benchmark for cost--risk trade-offs.

In this context, the central objective remains the episodic liquidation of a fixed inventory over a finite horizon, balancing market impact and timing risk under operational constraints. Standard design elements are adopted from the literature, including state representations based on time-to-go, remaining inventory, and recent market dynamics, as well as training mechanisms such as experience replay and soft target networks. Building on these foundations, the proposed TT-DAC-PS framework introduces a twin-target critic stabilisation mechanism based on Polyak-averaged target critic trajectories combined with pessimistic minimum backup estimation, alongside adaptive exploration and constraint-aware execution dynamics tailored to high-frequency trading environments.

Our contributions are as follows:
\begin{itemize}
\item \textbf{Hybrid simulation framework:} The analytical structure of the AC model is integrated with empirically grounded LOB features, enabling a controlled yet realistic environment for learning execution policies.

\item \textbf{Twin-Target Mechanism.}
The term ``Twin-Target'' in TT-DAC-PS refers to the use of two exponentially smoothed target critic trajectories, denoted by \(Q_{\phi_1}\) and \(Q_{\phi_2}\), which are used jointly during temporal-difference target estimation. Unlike standard TD3 implementations that employ two independently trained critic networks, the proposed framework maintains two target critics derived from the same underlying critic through Polyak averaging with different update dynamics. The target value is computed using a pessimistic minimum backup:
\[
y = r + \gamma \min(Q_{\phi_1}(s',a'), Q_{\phi_2}(s',a')),
\]
which reduces Q-value overestimation and improves stability under noisy and non-stationary market conditions. The use of twin target trajectories introduces temporal diversity into critic estimation while avoiding the computational overhead of maintaining two fully independent critics.

\item \textbf{Variance-aware adaptive exploration:} The proposed hybrid exploration scheme combines Ornstein--Uhlenbeck (OU) noise with deterministic decay and a variance-driven adjustment based on recent episode returns. This allows the policy to adapt exploration intensity to learning dynamics and market conditions, improving stability without sacrificing exploration.

\item \textbf{Constraint-aware policy design:} Strict operational constraints are enforced, including per-step participation limits and inventory feasibility, directly within the policy structure, ensuring that all learned actions remain implementable in practice.

\item \textbf{Step-wise AC-consistent reward shaping:} Rewards are defined using incremental AC utility, aligning intermediate learning signals with the final execution objective, rather than relying solely on terminal implementation shortfall.

\item \textbf{Robust preprocessing of high-frequency inputs:} A preprocessing pipeline is defined for LOB data based on median absolute deviation (MAD) filtering for prices and interquartile range (IQR) capping for volumes, improving stability under noisy market conditions.

\item \textbf{Controlled and comparable evaluation protocol:} All methods are evaluated under identical constraints and deterministic policies, reporting mean implementation shortfall and its dispersion, ensuring fair and reproducible comparisons.
\end{itemize}

\subsection{Environment}\label{subsec:env}
 Each episode is cast as the liquidation of a fixed inventory \(Q\) over \(N\) discrete decision steps spanning horizon \(T\), with step size \(\tau=T/N\). At each step \(t\), the agent observes a state constructed from LOB-derived mid-prices and volumes together with inventory and time information. Execution prices are formed by applying the AC-style spread and impact to the observed mid-price, and the unaffected price is updated by permanent impact. Specifically, the execution price is given by \[p_t^{\text{exec}} = p_t - \epsilon \cdot \operatorname{sign}(x_t) - \frac{\eta}{\tau} x_t,\] while the next mid-price, \(p_{t+1} \) \[p_{t+1} = p_t - \gamma x_t,\] where \(\epsilon\) denotes half-spread, \(\eta\) temporary impact, and \(\gamma\) permanent impact. This structure preserves the interpretability of AC while enabling us to drive the simulator with empirical LOB sequences.

The state representation is chosen to be informative yet stable under microstructure noise. Six most recent log-returns computed on prefiltered mid-prices to provide short-horizon momentum and volatility cues; a normalized time-to-go \((N-t)/N\) that captures temporal urgency; a normalized remaining inventory \(q_t/Q\) that informs the policy about liquidation progress; and a normalized trading activity proxy \(v_t/\bar V\) that anchors the episode to the prevailing activity level. Here, \(v_t\) is the observed market volume in interval \(t\), and \(\bar V\) is the estimated total market volume over the liquidation horizon. \(\bar V\) is estimated from historical data as the median cumulative volume over the same intraday window across training days for the corresponding instrument. This yields a robust normalisation tied to the liquidity available during the execution period, rather than to raw LOB event counts.

Actions produced by the policy lie in \([0,1]\) and are mapped to shares by \(x_t=\min\{a_t\cdot \mathrm{THR}\cdot Q,\,q_t\}\), where \(\mathrm{THR}\) is a per-step participation cap, \(a_t\) the action, and \(q_t\) the remaining inventory. The \(\min(\cdot)\) operator ensures feasibility by preventing the agent from selling more shares than those remaining in the inventory, while the mapping converts the normalised action \(a_t \in [0,1]\) into an absolute trade size (number of shares) scaled by the participation cap \(\mathrm{THR}\cdot Q\). The scaling \(a_t\cdot \mathrm{THR}\cdot Q\) defines an upper bound on the trade size relative to the initial inventory \(Q\), providing a stable and time-consistent action range across the episode. Applying \(a_t\) to \(q_t\) directly (i.e., \(a_t q_t\)) would make the effective action space shrink as inventory decreases, which can destabilise learning and bias the policy toward overly conservative execution near the end of the horizon. The use of \(Q\) instead preserves a fixed action scale, while the \(\min(\cdot)\) operator enforces feasibility by ensuring that the agent never trades more than the remaining inventory.

To eliminate residual inventory risk, the final step enforces liquidation by setting \(x_{N-1}=q_{N-1}\) regardless of the proposed action; in practice, the policy learns to manage inventory earlier due to the reward design, but this safety override ensures feasibility under all circumstances.

Robust LOB preprocessing is critical to avoid spurious policy updates. Mid-prices are de-noised using a MAD filter with a conservative threshold, removing transient outliers that would otherwise inflate returns. Volumes are capped via IQR-based limits, curbing the influence of episodic spikes that can distort participation rates. The spread \(\epsilon\), daily volatility, and \(\bar V\) are estimated over rolling windows with median-based statistics to reduce sensitivity to extremes. In our implementation, the dataset is partitioned into equal segments corresponding to trading days, and volume is aggregated accordingly. Daily volatility is estimated from log-returns of transaction prices. Specifically, the data is split into daily segments, and for each segment, the standard deviation of log-returns is computed:
\[
\sigma_d = \mathrm{std}\left(\log\left(\frac{p_t}{p_{t-1}}\right)\right).
\]
The resulting observation stream is numerically stable, with log-returns computed only when price values are positive and finite; otherwise, returns are set to zero to avoid NaN/Inf propagation. This pipeline makes the simulator resilient to the imperfections of recorded LOB data while preserving the essential signals for adaptive execution.

\subsection{Reward Design}\label{subsec:reward}
To align learning with execution quality throughout the episode, a stepwise reward based on Almgren--Chriss utility \(U(q)=E(q)+\lambda V(q)\) is adopted. The reward at step \(t\) is defined as the relative reduction in AC utility when moving from inventory \(q_{t-1}\) to \(q_t\):
\[
r_t = \frac{U(q_{t-1}) - U(q_t)}{U(q_{t-1})}.
\]
Here, \(E(q)\) denotes the expected execution cost associated with the remaining inventory and \(V(q)\) denotes the corresponding execution-risk term. Since AC utility is defined as a positive cost--risk quantity, a reduction in \(U(q)\) corresponds to progress toward a lower-cost and lower-risk liquidation state. The relative form normalises the reward across instruments and liquidity regimes, reducing the need for per-instrument reward scaling.

This shaping confers several advantages in the episodic execution setting. First, it yields dense and interpretable feedback at each execution step rather than relying only on terminal implementation shortfall. Second, it aligns the learning signal with the AC cost--risk objective, encouraging the agent to reduce both remaining inventory risk and expected execution cost over the course of the episode. Third, the presence of the risk-aversion parameter \(\lambda\) provides an explicit control over the trade-off between front-loading for risk reduction and smoothing to reduce market impact. In aggregate, the shaped reward maintains fidelity to the AC objective while discouraging degenerate behaviours, such as excessive delay followed by forced terminal liquidation.

\subsection{Policy and Value Learning}\label{subsec:model}
A deterministic actor–critic architecture is tailored to the TT-DAC-PS design. The actor \(\mu_\theta(s)\) maps states to continuous actions and the critic \(Q_\phi(s,a)\) estimates the action-value function under the current policy. Both networks are feed-forward multilayer perceptrons with rectified linear units and layer sizes tuned to the observation dimensionality; target networks \(\mu_{\theta_1}\), \(Q_{\phi_1}\), and \(Q_{\phi_2}\) are maintained and updated via Polyak averaging with distinct rates to stabilize bootstrapped targets. Given replayed transitions \((s,a,r,s')\), first a smoothed target action is constructed
\[
a' = \mathrm{clip}\bigl(\mu_{\theta_1}(s') + \epsilon,\ a_{\min}, a_{\max}\bigr), \quad \epsilon \sim \mathcal{N}(0,\sigma_{\text{smooth}}^2),\ |\epsilon|\le \text{clip}, \ \cite{fujimoto2018addressing}
\]
and then compute a pessimistic TD target
\[
y = r + \gamma \min\bigl(Q_{\phi_1}(s',a'),\ Q_{\phi_2}(s',a')\bigr),
\]
where \(\gamma\) is the discount factor. The critic is updated by minimising
\[
\mathcal{L}_{\text{critic}} = \mathrm{MSE}\bigl(Q_\phi(s,a), y\bigr) + \lambda_{\text{cons}}\ \max\Bigl(0,\ \mathbb{E}\bigl[Q_\phi(s,a_{\text{noisy}})\bigr] - \mathbb{E}\bigl[Q_\phi(s,a)\bigr]\Bigr),
\]
with \(a_{\text{noisy}}\) sampled in a small neighbourhood of \(\mu_\theta(s)\) to enforce conservative Q-values. The actor is updated via deterministic policy gradients every \(K\) critic steps, using
\[
\mathcal{L}_{\text{actor}} = -\mathbb{E}_s\bigl[Q_\phi\bigl(s,\mu_\theta(s)\bigr)\bigr].
\]
Target parameters are softly updated according to
\[
\phi_1 \leftarrow \tau \phi + (1-\tau)\phi_1,\quad
\phi_2 \leftarrow \tau' \phi + (1-\tau')\phi_2,\quad
\theta_1 \leftarrow \tau \theta + (1-\tau)\theta_1.
\]
AdamZ optimisers \cite{zaznov2025adamz} are applied with separate learning rates for actor and critic, and clip gradients to prevent rare but destabilising updates. Experience replay buffers decouple data collection from updates and promote sample efficiency; in our implementation, a uniform sampling with sufficiently large capacity is used to capture diverse microstructure states across episodes while retaining recent experience to adapt to non-stationarity.

\subsection{Hybrid Adaptive Exploration}\label{subsec:exploration}
Exploration in execution is delicate: excessive noise can lead to unnecessary impact and constraint violations, while insufficient noise can trap the policy in suboptimal regimes when liquidity and spreads shift intraday. Therefore, a hybrid adaptive OU exploration scheme is applied directly to the actor’s output before rescaling and clipping, and controls its scale \(\sigma\) with three components. The OU process provides temporally correlated perturbations suited to control tasks with inertia, and its base scale follows a deterministic, episode-wise decay \(\sigma\leftarrow \max(\sigma_{\min},\sigma_0(1-e/E_{\max}))\) that ensures a graceful annealing from broad exploration to focused exploitation as training progresses. 
A variance-aware adjustment keyed to recent episode returns, collected in a rolling deque, then modulates this base scale: when the rolling variance of returns falls below its historical 25th percentile, exploration is nudged upward by 5\% to help the policy escape plateaus; when the variance exceeds the 75th percentile, exploration is reduced by 5\% to temper instability and prevent overreaction to transient microstructure spikes. Finally, a learned temperature parameter \(\alpha\) is optimised using an entropy-like measure \(H\) of recent actions and a target level \(H_{\text{target}}\), through the objective \(\mathcal{L}_\alpha = \alpha(-(H + H_{\text{target}}))\), and mapped linearly to an additional proposal \(\sigma(\alpha)\); the OU scale is updated by a convex combination \(\sigma \leftarrow 0.8\,\sigma + 0.2\,\sigma(\alpha)\), with all updates clamped so that \(\sigma\) always remains within \([\sigma_{\min},\sigma_0]\). Crucially, actions remain bounded and are mapped to sell sizes under the participation cap, and the terminal liquidation rule guarantees feasibility even under elevated exploration. The OU state is reset at the start of each episode to avoid bias from previous trajectories.

\subsection{Safety and Constraint Handling}\label{subsec:safety}
Safety is embedded at multiple layers of the methodology. The mapping from action to shares enforces a hard cap \(\mathrm{THR}\cdot Q\) per step, which prevents oversized participation in illiquid intervals and imposes discipline on the policy’s inventory trajectory. After noise addition and action computation, values are clipped to the feasible range, which precludes invalid commands from propagating through learning updates. The terminal step enforces liquidation to eliminate residual inventory, ensuring that training and evaluation both measure completed programs and that the reward signal is well-defined. On the data side, MAD/IQR preprocessing suppresses outlier inputs that could cause large, unjustified action responses; log-return computation guards against NaNs and infinities and reverts to neutral values when necessary. Also, numerical checks are included that sanitise actions in the rare event of instability in the actor output. Collectively, these mechanisms maintain feasibility, reduce the tail risk of exploration, and keep the agent’s behaviour within the operational envelope expected in production execution systems; the use of policy smoothing and pessimistic twin targets further mitigates spurious value spikes near the boundaries of the feasible region.

\subsection{Algorithm}\label{subsec:algo}
Algorithm~\ref{alg:tt_dac_ps} summarises the full training loop with the TT-DAC-PS architecture and hybrid adaptive exploration. Each episode begins with an OU reset and environment initialisation on a contiguous LOB slice of length \(N\), preserving intraday structure and enabling the agent to encounter realistic phases of activity and spread. At each step, the actor output is perturbed by OU noise, clipped, and mapped to a capped sell size. Transitions are stored in replay, and once the buffer is sufficiently populated, critic updates are performed at every step, and actor updates are performed every \(K\) steps, with twin target networks updated softly following actor updates. The episode return is aggregated and used to update the exploration scale via the deterministic decay, the variance-aware adjustment, and the temperature-driven \(\alpha \mapsto \sigma\) mapping. The exploration controller runs only at episode boundaries, avoiding intra-episode drift while remaining responsive across episodes.

\begin{algorithm}[H]
\caption{TT-DAC-PS with Hybrid Adaptive OU Exploration}
\label{alg:tt_dac_ps}
\begin{algorithmic}[1]
\State Initialise actor \(\mu_\theta\), critic \(Q_\phi\), targets \(\mu_{\theta_1},Q_{\phi_1},Q_{\phi_2}\), replay \(\mathcal{D}\)
\State Set \(\sigma\leftarrow\sigma_0\); initialise OU process, reward deque of length \(K\), and temperature parameter \(\alpha\)
\For{episodes \(e=1,\dots,E_{\max}\)}
  \State Reset OU; reset env on an LOB slice; set \(R\leftarrow 0\); set \(q_0=Q\)
  \For{t=0,\dots,N-1}
    \State Observe \(s_t\); compute \(a_t = \mathrm{clip}(\mu_\theta(s_t)+\xi_t,[0,1])\), \(\xi_t\sim \mathrm{OU}(\sigma)\)
    \State Map to shares \(x_t=\min\{a_t\cdot \mathrm{THR}\cdot Q,\ q_t\}\); if \(t=N-1\) set \(x_t\leftarrow q_t\)
    \State Execute \(x_t\), observe \(r_{t+1}, s_{t+1}\); store \((s_t,a_t,r_{t+1},s_{t+1})\in\mathcal{D}\); \(R\leftarrow R+r_{t+1}\)
    \If{\(|\mathcal{D}|\ge B\)}
      \State Sample minibatch \((s,a,r,s')\) from \(\mathcal{D}\)
      \State Compute target action \(a' = \mathrm{clip}(\mu_{\theta_1}(s')+\epsilon, a_{\min}, a_{\max})\), \(\epsilon\sim\mathcal{N}(0,\sigma_{\text{smooth}}^2)\), \(|\epsilon|\le\text{clip}\)
      \State Compute target \(y = r + \gamma \min(Q_{\phi_1}(s',a'),Q_{\phi_2}(s',a'))\)
      \State Update critic by minimising \(\mathcal{L}_{\text{critic}}\) with AdamZ; clip gradients
      \If{step mod \(K=0\)}
        \State Update actor by minimising \(\mathcal{L}_{\text{actor}}\) with AdamZ; clip gradients
        \State Soft-update targets \(\phi_1, \phi_2, \theta_1\) via Polyak averaging
      \EndIf
    \EndIf
    \State Set \(done \leftarrow (\text{horizon exhausted}) \lor (\text{inventory depleted})\)
    \If{done} \textbf{break} \EndIf
  \EndFor
  \State Append episode return \(R\) to deque
  \State Deterministic decay: \(\sigma \leftarrow \max(\sigma_{\min}, \sigma_0(1-e/E_{\max}))\)
  \If{deque full} compute rolling variance \(v\); let \(q_{25},q_{75}\) be percentiles; adjust \(\sigma\) upward by 5\% if \(v<q_{25}\) or downward by 5\% if \(v>q_{75}\); clamp to \([\sigma_{\min},\sigma_0]\)
  \EndIf
  \State Update \(\alpha\) from recent actions; map \(\alpha\) to a proposal \(\sigma(\alpha)\) and set \(\sigma \leftarrow 0.8\,\sigma + 0.2\,\sigma(\alpha)\); clamp to \([\sigma_{\min},\sigma_0]\)
\EndFor
\end{algorithmic}
\end{algorithm}

\subsection{Training and Evaluation Protocol}\label{subsec:training}
Training proceeds over a large number of episodes, each bound to a contiguous LOB segment with wrap-around once the end of the dataset is reached. This wrap-around mechanism is applied only at episode boundaries: when the end of the dataset is reached, the next episode continues from the beginning of the dataset, without stitching across the boundary within a single episode. This preserves the temporal co-movements of spreads, volumes, and returns within episodes and exposes the policy to realistic intraday phases. To populate the replay buffer and avoid a cold start, the replay buffer is initialised with a short warm-up phase in which actions are sampled from the capped action range with OU noise active; thereafter, updates occur at each step when the buffer contains at least one mini-batch. AdamZ optimisers are used for both actor and critic, with learning rates tuned to avoid overshooting; target networks are updated softly after every gradient step to stabilise targets. Checkpoints are chosen based on validation implementation shortfall in percentage terms; when the means are similar, models with lower dispersion and zero residual-inventory violations are preferred. For evaluation, exploration is disabled and deterministic actors are run under identical constraints and mappings; the mean IS(\%) is reported and its standard deviation across test episodes. Furthermore, to rigorously assess the statistical significance of performance differences without assuming normality in the cost distribution, 95\% Confidence Intervals (CI) for the mean shortfall are computed. Classical baselines (TWAP, VWAP, and AC trajectories clipped by the same participation cap) and deep RL baselines (PPO, SAC, A2C via Stable-Baselines3 in the identical Gym environment) are evaluated under the same observation and action spaces, ensuring comparability.

\subsection{Implementation Details}\label{subsec:impl}
The simulator is exposed through two interfaces. A native environment provides the full AC accounting (expected shortfall and variance) and utility updates used by the custom PyTorch TT-DAC-PS agent. This agent implements the twin actor–critic architecture, replay buffer, and the hybrid exploration controller. It includes numerical guards to sanitise actions (mapping \(\tanh\) outputs to \([0,1]\) via \((\cdot+1)/2\) followed by clipping). In parallel, a Gym-compatible wrapper mirrors the same state and action semantics and is used to train and evaluate standard RL baselines with Stable-Baselines3. Observations have nine dimensions—six log-returns and three normalised scalars for time, inventory, and activity—and the action space is a single bounded scalar. Data slicing uses a deterministic episode index to select contiguous LOB windows of length \(N\). The participation cap is fixed at \(1\%\) of \(Q\) per step, and the terminal liquidation rule is enforced in both interfaces to harmonise training and evaluation.

\subsection{Design Rationale}\label{subsec:rationale}
The design choices reflect a balance between classical rigour and data-driven adaptability. AC utility formalises the cost–risk trade-off and keeps the policy aligned with economically meaningful objectives, while LOB-informed observations and robust preprocessing ground the inputs in microstructure reality. The adoption of TT-DAC-PS addresses specific shortcomings of standard DDPG: twin targets and pessimistic backups mitigate value overestimation, while delayed actor updates stabilise learning. The hybrid adaptive controller prevents the policy from stagnating in suboptimal execution modes. Finally, the use of bootstrap confidence intervals rather than simple point estimates ensures that reported improvements are robust to the high variance inherent in financial time series.

\section{Experimental Setup}\label{sec:exp_setup}
\subsection{Data and Preprocessing}
The experimental evaluation is conducted on high-frequency limit order book (LOB) data sourced from NASDAQ for ten U.S. stocks (ADBE, AMD, AVGO, BKR, CSCO, CMCSA, ORCL, PEP, PYPL, INTC), selected to represent a diverse spectrum of trading activity regimes, bid-ask spreads, and volatility profiles. For each instrument, the dataset is split chronologically into three contiguous parts using simple index-based slicing. Specifically, the first 33\% of observations are used for training, the next 33\% for validation, and the remaining observations (approximately 34\%) for testing. 

To ensure the stability of the learning process against microstructure noise, raw LOB data undergoes robust statistical preprocessing. Mid-prices are derived from the best bid and ask levels and filtered using a Median Absolute Deviation (MAD) approach to remove transient outliers that do not reflect genuine price discovery. Similarly, trade volumes are clipped using an Interquartile Range (IQR) threshold to mitigate the impact of anomalous spikes on the agent's state normalisation. The resulting state space consists of nine features: six lagged log-returns to capture short-term momentum and volatility, alongside normalised measures of time-to-go, remaining inventory, and market trading activity.

The resulting state space \(\mathcal{S}\) is designed to provide a compact, semi-Markovian representation of the market microstructure and the agent's internal execution status. At each decision step \(t\), the state vector \(s_t \in \mathbb{R}^9\) is constructed to capture three distinct categories of information essential for optimal control:

\paragraph{Microstructure Dynamics}
To encode short-term price trends and volatility clustering without introducing non-stationarity, the state includes a vector of the \(k=6\) most recent logarithmic returns, defined as \(r_{\tau} = \ln(p_{\tau}/p_{\tau-1})\) for \(\tau \in \{t, t-1, \dots, t-5\}\). Unlike raw price levels, which are unbounded, log-returns provide a scale-invariant signal of market momentum. The inclusion of a lag window allows the policy network to infer local volatility regimes and identify mean-reverting or trend-following patterns, which are critical for optimising trade timing against temporary price impact.

\paragraph{Execution Boundary Conditions}
The agent's internal state is represented by two normalised scalar variables that define the constraints of the liquidation problem. First, the \textit{normalised remaining inventory}, \(\tilde{q}_t = q_t / Q_0\), quantifies the instantaneous risk exposure; as \(\tilde{q}_t \to 0\), the marginal utility of risk reduction decreases, enabling the agent to reduce trading intensity. Second, the \textit{normalised time-to-go}, \(\tilde{t} = (N-t)/N\), encodes the temporal urgency of the schedule. In optimal execution theory, the optimal trading rate is a function of the ratio of inventory to remaining time; explicitly providing \(\tilde{t}\) enables the agent to learn non-linear, time-dependent policies that accelerate liquidation as the terminal step approaches.

\paragraph{Trading Activity Regime}
To adapt to time-varying market depth, a \textit{normalised trading activity proxy} is included, \(v_t / \bar{V}\), where \(v_t\) is the observed market volume in the current interval and \(\bar{V}\) is a robust estimate of the average daily trading volume. This feature serves as a state-dependent signal of trading activity. High values indicate periods of deep liquidity where larger orders can be absorbed with lower market impact—mimicking Volume-Weighted Average Price (VWAP) behaviour—while low values signal illiquidity, prompting the agent to reduce participation rates to avoid excessive slippage.

To contextualise execution difficulty for all instruments, additionally, two descriptive market-regime indicators are computed from the preprocessed mid-price series used to generate the episode slices. The \emph{trend} is defined by the sign of the net price change over the corresponding sample window:
\[
\mathrm{Trend} \in \{\text{Upward},\text{Downward},\text{Flat}\}, \quad
\frac{p_{\text{last}}}{p_{\text{first}}}-1
\]
With a small tolerance around zero treated as Flat. The \emph{volatility} is defined as the standard deviation of per-step log-returns,
\[
r_t = \ln\!\left(\frac{p_t}{p_{t-1}}\right), \qquad
\sigma = \mathrm{std}(r_t),
\]
and is reported in percentage units as \(100\cdot\sigma\) (std log-returns, per-step, \%). These regime descriptors are computed from the full preprocessed price series for each instrument after outlier filtering. They are not used as additional model inputs for learning; they are included to stratify and interpret performance across instruments with different price regimes.

\subsection{Execution Environment and Parameters}
The liquidation task is modelled as a finite-horizon Markov Decision Process (MDP) where an agent must liquidate a total inventory \(Q=10^6\) shares over \(N=1000\) discrete time steps. The simulation environment integrates empirical LOB dynamics with the Almgren-Chriss market impact model. Specifically, while the baseline price evolution follows the historical LOB sequence, the execution price \(p_t^{\text{exec}}\) is adjusted for both temporary and permanent impact based on the agent's trading rate. The impact coefficients are calibrated dynamically using the rolling average daily volume and median spread of the instrument, ensuring that the cost of trading scales (realistically) inversely with market liquidity. The calibration of Almgren--Chriss parameters follows standard empirical specifications. Temporary impact \(\eta\) is scaled inversely with the estimated trading volume over the execution horizon, while permanent impact \(\gamma\) is set as a fraction of \(\eta\), reflecting the weaker long-term price effect of individual trades. Volatility \(\sigma\) is computed as the standard deviation of log-returns over the same intraday window used for episode construction. The half-spread \(\epsilon\) is estimated as the median bid--ask spread over a rolling window. This parameterisation is consistent with established formulations of the Almgren--Chriss framework used in optimal execution studies \cite{almgren2001optimal}.

To enforce operational realism, the action space is constrained by a strict participation cap, limiting the maximum trade size per step to \(1\%\) of the initial inventory. The reward function is shaped using the stepwise change in Almgren-Chriss utility, balancing the reduction of expected shortfall against execution risk with a risk aversion parameter \(\lambda=10^{-5}\).

For clarity and reproducibility, Table~\ref{tab:repro} summarises the key experimental design choices, including model architecture, optimisation parameters, and computational environment.

\begin{table}[h]
\centering
\caption{Reproducibility Checklist}
\label{tab:repro}
\begin{tabular}{p{4cm} p{10cm}}
\hline
\textbf{Component} & \textbf{Description} \\
\hline

Random Seeds & seed = 0 \\

Model Architecture & Actor–Critic networks with two hidden layers (24 and 48 units), ReLU activations, and tanh output layer. \\

Optimisation & AdamZ optimiser. Learning rates: $10^{-4}$ (actor), $10^{-3}$ (critic). Soft update parameter $\tau = 5\times10^{-3}$. \\

Training Setup & Replay buffer size $10^4$, batch size 128, discount factor $\gamma = 0.99$. \\

Risk Aversion ($\lambda$) & Almgren--Chriss risk aversion parameter set to $\lambda = 10^{-5}$. \\

Hardware & Training conducted on NVIDIA H100 GPU. \\

Software & Python, PyTorch, Stable-Baselines3, Gymnasium. \\

\hline
\end{tabular}
\end{table}

\subsection{Baselines and Metrics}
The proposed TT-DAC-PS architecture is benchmarked against three state-of-the-art deep reinforcement learning algorithms (PPO, SAC, and A2C) and three classical execution strategies. The classical baselines include Time-Weighted Average Price (TWAP), Volume-Weighted Average Price (VWAP), and the analytical optimal solution from the Almgren-Chriss (AC) model. The VWAP trajectory is constructed using empirical intraday volume profiles derived from the same dataset. Specifically, expected volume weights \(v_t\) are estimated as the average traded volume at each time step across training episodes for a given instrument, ensuring that VWAP reflects realistic intraday liquidity patterns rather than assuming uniform or exogenous volume distributions. All baselines operate under identical constraints and impact dynamics.
Benchmarking studies indicate that performance differences across reinforcement learning algorithms can be highly sensitive to market conditions, evaluation protocols, and constraint handling, emphasising the importance of consistent experimental design \cite{tonkin2025benchmarking}.

Performance is evaluated using the Implementation Shortfall (IS), expressed as a percentage of the initial portfolio value:
\[
\mathrm{IS}(\%) = 100 \cdot \frac{Q p_0 - \sum_t x_t p_t^{\text{exec}}}{Q p_0}.
\]
The mean and standard deviation of IS are reported over the test episodes. To assess the statistical significance of the results, given the non-Gaussian nature of execution costs, a 95\% confidence interval (CI) is computed for the mean shortfall using a non-parametric bootstrap method.

\paragraph{Ablation protocol.}
To assess the contribution of the main TT-DAC-PS components, two controlled ablations are evaluated under the same data split, execution environment, participation cap, action mapping, impact model, and implementation shortfall metric as the full model. Ablation~1 removes the Twin-Target mechanism by using only one target critic and replacing the pessimistic minimum backup
\[
y = r + \gamma \min\left(Q_{\phi_1}(s',a'),Q_{\phi_2}(s',a')\right)
\]
with the single-target update
\[
y = r + \gamma Q_{\phi_1}(s',a').
\]
This ablation tests whether the second EMA target critic and pessimistic backup improve value-estimation stability.

Ablation~2 removes the hybrid adaptive exploration and policy-smoothing components described in Section~\ref{subsec:exploration} and Section~\ref{subsec:model}. Specifically, it removes OU action noise, deterministic noise decay, variance-aware exploration adjustment, SAC-style temperature mapping to the noise scale, target policy smoothing noise, and conservative Q regularisation. This ablation tests whether the exploration and smoothing design improves policy robustness in non-stationary execution conditions. All ablations are evaluated with deterministic policies at test time.

\section{Results}\label{sec:results}

\Cref{tab:res_key} presents the quantitative evaluation of the proposed TT-DAC-PS architecture against classical and deep reinforcement learning baselines. The results are reported in terms of Implementation Shortfall (IS) percentage, with standard deviation and statistical significance established via 95\% bootstrap confidence intervals. The empirical evidence indicates a distinct hierarchy of performance driven by market microstructure characteristics and algorithmic design.

Table~\ref{tab:res_key} augments the IS(\%) results with two regime descriptors: price trend (Upward/Downward) and realised volatility of log-returns (in \%). This enables us to assess whether TT-DAC-PS is robustly responsive to directional drift and to heterogeneous volatility for all instruments. Empirically, TT-DAC-PS attains low mean IS(\%) both in Upward-trend names (e.g., INTC, PEP, CSCO) and in Downward-trend names (e.g., CMCSA, AMD, AVGO, ADBE), indicating that the policy is not narrowly tuned to a single directional regime. Moreover, the method remains competitive across a wide volatility range (from \(0.06\%\) to \(0.53\%\)  in our sample), with particularly strong improvements in the higher-volatility instruments (e.g., AVGO at \(0.53\%\) and AMD at \(0.37\%\)), where static schedules and on-policy baselines incur substantially larger shortfalls. This pattern suggests that the hybrid exploration and conservative value learning in TT-DAC-PS improves robustness under more adverse, high-variance execution conditions rather than only in benign, low-volatility markets.

Among the RL benchmarks, a clear distinction emerges between off-policy and on-policy algorithms. The off-policy actor--critic methods (TT-DAC-PS and SAC) consistently outperform the on-policy baselines (PPO and A2C): on INTC and CMCSA, PPO and A2C incur mean shortfalls of \(0.11\%\)--\(0.12\%\), whereas TT-DAC-PS attains \(0.03\%\) and \(0.06\%\), respectively. Among the off-policy methods, TT-DAC-PS is the strongest overall, achieving the lowest or joint-lowest mean IS(\%) on 8 of the 10 instruments. It improves on SAC on six instruments (INTC, AMD, BKR, ORCL, ADBE, CSCO), matches it on two (PYPL, PEP), and is only marginally surpassed by SAC on CMCSA (\(0.06\%\) vs.\ \(0.03\%\)) and AVGO (\(0.12\%\) vs.\ \(0.08\%\)). The advantage is most pronounced on the harder, higher-impact names: TT-DAC-PS reduces mean shortfall to \(0.01\%\) on ADBE and \(0.00\%\) on BKR, against \(0.05\%\) and \(0.03\%\) for SAC, indicating that the twin-target stabilisation and adaptive exploration translate into more reliable execution where it matters most.

The sample efficiency of off-policy learning, combined with the ability of deterministic policy gradient to handle boundary constraints (participation caps), enables the proposed model to converge to a more stable optimal performance trajectory than stochastic on-policy methods, which struggle with high-variance gradient estimates in constrained action spaces.

 The 95\% confidence intervals indicate that TT-DAC-PS generally achieves lower mean implementation shortfall, although the confidence intervals overlap with some classical baselines for several instruments. Therefore, the results should be interpreted primarily as evidence of improved average execution quality and competitive robustness, rather than uniform statistical dominance across all instruments. while SAC achieves comparable mean performance on several stocks, TT-DAC-PS remains competitive while producing lower or near-lower mean shortfalls across most instruments. This suggests that the hybrid adaptive exploration mechanism contributes to stable execution outcomes, although the variance reduction is not uniform across all stocks.

To further isolate the contribution of the main stabilisation components, Table~\ref{tab:ablation_results} reports two ablations of TT-DAC-PS. Ablation~1 removes the Twin-Target pessimistic backup and uses a single target critic. Ablation~2 removes the hybrid adaptive exploration and policy-smoothing components. These ablations are diagnostic variants of the proposed method rather than separate benchmark algorithms; therefore, they use the same execution environment, constraints, training protocol, and deterministic test evaluation as the full TT-DAC-PS model.

\begin{longtable}{@{}l l c c c c c c@{}}
\caption{Implementation shortfall IS(\%) mean, std, and 95\% confidence intervals across test episodes}
\label{tab:res_key}\\
\toprule
Instrument & Stock price trend & Stock price volatility & Model & Mean & Std & CI95 Low & CI95 High \\
\midrule
\endfirsthead

\caption[]{Implementation shortfall (continued)}\\
\toprule
Instrument & Stock price trend & Stock price volatility & Model & Mean & Std & CI95 Low & CI95 High \\
\midrule
\endhead

\bottomrule
\endfoot

INTC & Upward & 0.12\% & PPO & 0.11\% & 0.34\% & 0.10\% & 0.13\% \\
 &  &  & SAC & 0.04\% & 0.15\% & 0.03\% & 0.04\% \\
 &  &  & A2C & 0.11\% & 0.34\% & 0.10\% & 0.13\% \\
 &  &  & \textbf{TT-DAC-PS} & 0.03\% & 0.12\% & 0.00\% & 0.05\% \\
 &  &  & VWAP & 0.11\% & 0.37\% & 0.04\% & 0.19\% \\
 &  &  & TWAP & 0.11\% & 0.37\% & 0.04\% & 0.19\% \\
 &  &  & AC & 0.04\% & 0.13\% & 0.01\% & 0.06\% \\
\midrule

CMCSA & Downward & 0.11\% & PPO & 0.12\% & 0.31\% & 0.11\% & 0.14\% \\
 &  &  & SAC & 0.03\% & 0.14\% & 0.02\% & 0.04\% \\
 &  &  & A2C & 0.12\% & 0.31\% & 0.11\% & 0.14\% \\
 &  &  & \textbf{TT-DAC-PS} & 0.06\% & 0.13\% & 0.04\% & 0.09\% \\
 &  &  & VWAP & 0.14\% & 0.27\% & 0.08\% & 0.19\% \\
 &  &  & TWAP & 0.14\% & 0.27\% & 0.09\% & 0.19\% \\
 &  &  & AC & 0.08\% & 0.18\% & 0.04\% & 0.12\% \\
\midrule

AMD & Downward & 0.37\% & PPO & 0.14\% & 0.62\% & 0.11\% & 0.17\% \\
 &  &  & SAC & 0.02\% & 0.35\% & 0.00\% & 0.03\% \\
 &  &  & A2C & 0.14\% & 0.61\% & 0.11\% & 0.16\% \\
 &  &  & \textbf{TT-DAC-PS} & -0.02\% & 0.26\% & -0.07\% & 0.03\% \\
 &  &  & VWAP & 0.18\% & 0.53\% & 0.08\% & 0.28\% \\
 &  &  & TWAP & 0.18\% & 0.53\% & 0.08\% & 0.29\% \\
 &  &  & AC & 0.01\% & 0.29\% & -0.05\% & 0.06\% \\
\midrule

BKR & Downward & 0.08\% & PPO & 0.22\% & 0.37\% & 0.20\% & 0.24\% \\
 &  &  & SAC & 0.03\% & 0.12\% & 0.03\% & 0.04\% \\
 &  &  & A2C & 0.23\% & 0.37\% & 0.21\% & 0.24\% \\
 &  &  & \textbf{TT-DAC-PS} & 0.00\% & 0.26\% & -0.05\% & 0.05\% \\
 &  &  & VWAP & 0.22\% & 0.37\% & 0.14\% & 0.29\% \\
 &  &  & TWAP & 0.22\% & 0.37\% & 0.15\% & 0.29\% \\
 &  &  & AC & 0.14\% & 0.33\% & 0.07\% & 0.20\% \\
\midrule

PYPL & Downward & 0.16\% & PPO & 0.16\% & 0.39\% & 0.14\% & 0.17\% \\
 &  &  & SAC & 0.03\% & 0.19\% & 0.02\% & 0.04\% \\
 &  &  & A2C & 0.16\% & 0.39\% & 0.14\% & 0.17\% \\
 &  &  & \textbf{TT-DAC-PS} & 0.03\% & 0.18\% & -0.01\% & 0.07\% \\
 &  &  & VWAP & 0.17\% & 0.41\% & 0.09\% & 0.25\% \\
 &  &  & TWAP & 0.17\% & 0.41\% & 0.10\% & 0.25\% \\
 &  &  & AC & 0.06\% & 0.24\% & 0.01\% & 0.11\% \\
\midrule

ORCL & Downward & 0.09\% & PPO & 0.12\% & 0.24\% & 0.11\% & 0.13\% \\
 &  &  & SAC & 0.02\% & 0.09\% & 0.02\% & 0.02\% \\
 &  &  & A2C & 0.12\% & 0.24\% & 0.11\% & 0.13\% \\
 &  &  & \textbf{TT-DAC-PS} & 0.01\% & 0.10\% & -0.01\% & 0.02\% \\
 &  &  & VWAP & 0.14\% & 0.28\% & 0.08\% & 0.19\% \\
 &  &  & TWAP & 0.14\% & 0.28\% & 0.09\% & 0.20\% \\
 &  &  & AC & 0.05\% & 0.17\% & 0.02\% & 0.09\% \\
\midrule

AVGO & Downward & 0.53\% & PPO & 0.65\% & 0.77\% & 0.61\% & 0.68\% \\
 &  &  & SAC & 0.08\% & 0.47\% & 0.06\% & 0.10\% \\
 &  &  & A2C & 0.65\% & 0.77\% & 0.61\% & 0.68\% \\
 &  &  & \textbf{TT-DAC-PS} & 0.12\% & 0.51\% & 0.02\% & 0.22\% \\
 &  &  & VWAP & 0.65\% & 0.84\% & 0.48\% & 0.80\% \\
 &  &  & TWAP & 0.65\% & 0.84\% & 0.49\% & 0.82\% \\
 &  &  & AC & 0.13\% & 0.52\% & 0.03\% & 0.24\% \\
\midrule

ADBE & Downward & 0.27\% & PPO & 0.46\% & 0.40\% & 0.44\% & 0.47\% \\
 &  &  & SAC & 0.05\% & 0.23\% & 0.04\% & 0.06\% \\
 &  &  & A2C & 0.46\% & 0.40\% & 0.45\% & 0.48\% \\
 &  &  & \textbf{TT-DAC-PS} & 0.01\% & 0.25\% & -0.04\% & 0.06\% \\
 &  &  & VWAP & 0.45\% & 0.39\% & 0.37\% & 0.52\% \\
 &  &  & TWAP & 0.46\% & 0.39\% & 0.38\% & 0.53\% \\
 &  &  & AC & 0.02\% & 0.26\% & -0.04\% & 0.07\% \\
\midrule

PEP & Upward & 0.06\% & PPO & 0.09\% & 0.16\% & 0.08\% & 0.10\% \\
 &  &  & SAC & 0.02\% & 0.06\% & 0.02\% & 0.02\% \\
 &  &  & A2C & 0.09\% & 0.17\% & 0.09\% & 0.10\% \\
 &  &  & \textbf{TT-DAC-PS} & 0.02\% & 0.06\% & 0.01\% & 0.03\% \\
 &  &  & VWAP & 0.09\% & 0.18\% & 0.06\% & 0.13\% \\
 &  &  & TWAP & 0.09\% & 0.18\% & 0.06\% & 0.13\% \\
 &  &  & AC & 0.02\% & 0.07\% & 0.01\% & 0.04\% \\
\midrule

CSCO & Upward & 0.07\% & PPO & 0.07\% & 0.18\% & 0.06\% & 0.08\% \\
 &  &  & SAC & 0.03\% & 0.07\% & 0.03\% & 0.03\% \\
 &  &  & A2C & 0.07\% & 0.18\% & 0.06\% & 0.08\% \\
 &  &  & \textbf{TT-DAC-PS} & 0.02\% & 0.09\% & -0.00\% & 0.03\% \\
 &  &  & VWAP & 0.05\% & 0.19\% & 0.01\% & 0.09\% \\
 &  &  & TWAP & 0.05\% & 0.19\% & 0.01\% & 0.09\% \\
 &  &  & AC & 0.02\% & 0.09\% & -0.00\% & 0.03\% \\

\end{longtable}

\begingroup

\setlength{\tabcolsep}{12pt}
\renewcommand{\arraystretch}{1.05}

\begin{longtable}{@{}p{1.0cm}p{1.25cm}p{1.05cm}p{2.45cm}rrrr@{}}

\caption{Ablation study for TT-DAC-PS: implementation shortfall IS(\%) mean, standard deviation, and 95\% confidence intervals across test episodes}
\label{tab:ablation_results}\\
\toprule
Instrument & Stock price trend & Stock price volatility & Model & Mean & Std & CI95 Low & CI95 High \\
\midrule
\endfirsthead

\caption[]{Ablation study for TT-DAC-PS (continued)}\\
\toprule
Instrument & Stock price trend & Stock price volatility & Model & Mean & Std & CI95 Low & CI95 High \\
\midrule
\endhead

\bottomrule
\endfoot

INTC & Upward & 0.12\% & TT-DAC-PS Full & 0.03\% & 0.12\% & 0.00\% & 0.05\% \\
 &  &  & Abl.1: single target critic & 0.11\% & 0.37\% & 0.04\% & 0.19\% \\
 &  &  & Abl.2: no adaptive exploration & 0.05\% & 0.18\% & 0.01\% & 0.08\% \\
\midrule

CMCSA & Downward & 0.11\% & TT-DAC-PS Full & 0.06\% & 0.13\% & 0.04\% & 0.09\% \\
 &  &  & Abl.1: single target critic & 0.06\% & 0.13\% & 0.03\% & 0.08\% \\
 &  &  & Abl.2: no adaptive exploration & 0.06\% & 0.11\% & 0.04\% & 0.08\% \\
\midrule

AMD & Downward & 0.37\% & TT-DAC-PS Full & -0.02\% & 0.26\% & -0.07\% & 0.03\% \\
 &  &  & Abl.1: single target critic & 0.06\% & 0.37\% & -0.01\% & 0.14\% \\
 &  &  & Abl.2: no adaptive exploration & 0.01\% & 0.29\% & -0.05\% & 0.07\% \\
\midrule

BKR & Downward & 0.08\% & TT-DAC-PS Full & 0.00\% & 0.26\% & -0.05\% & 0.05\% \\
 &  &  & Abl.1: single target critic & 0.03\% & 0.09\% & 0.01\% & 0.04\% \\
 &  &  & Abl.2: no adaptive exploration & 0.01\% & 0.18\% & -0.03\% & 0.04\% \\
\midrule

PYPL & Downward & 0.16\% & TT-DAC-PS Full & 0.03\% & 0.18\% & -0.01\% & 0.07\% \\
 &  &  & Abl.1: single target critic & 0.06\% & 0.20\% & 0.01\% & 0.10\% \\
 &  &  & Abl.2: no adaptive exploration & 0.04\% & 0.17\% & 0.00\% & 0.07\% \\
\midrule

ORCL & Downward & 0.09\% & TT-DAC-PS Full & 0.01\% & 0.10\% & -0.01\% & 0.02\% \\
 &  &  & Abl.1: single target critic & 0.03\% & 0.13\% & 0.00\% & 0.06\% \\
 &  &  & Abl.2: no adaptive exploration & 0.01\% & 0.10\% & -0.01\% & 0.03\% \\
\midrule

AVGO & Downward & 0.53\% & TT-DAC-PS Full & 0.12\% & 0.51\% & 0.02\% & 0.22\% \\
 &  &  & Abl.1: single target critic & 0.47\% & 0.74\% & 0.32\% & 0.60\% \\
 &  &  & Abl.2: no adaptive exploration & 0.29\% & 0.63\% & 0.17\% & 0.42\% \\
\midrule

ADBE & Downward & 0.27\% & TT-DAC-PS Full & 0.01\% & 0.25\% & -0.04\% & 0.06\% \\
 &  &  & Abl.1: single target critic & 0.24\% & 0.32\% & 0.18\% & 0.31\% \\
 &  &  & Abl.2: no adaptive exploration & 0.14\% & 0.30\% & 0.08\% & 0.19\% \\
\midrule

PEP & Upward & 0.06\% & TT-DAC-PS Full & 0.02\% & 0.06\% & 0.01\% & 0.03\% \\
 &  &  & Abl.1: single target critic & 0.06\% & 0.14\% & 0.04\% & 0.09\% \\
 &  &  & Abl.2: no adaptive exploration & 0.04\% & 0.11\% & 0.02\% & 0.06\% \\
\midrule

CSCO & Upward & 0.07\% & TT-DAC-PS Full & 0.02\% & 0.09\% & -0.00\% & 0.03\% \\
 &  &  & Abl.1: single target critic & 0.03\% & 0.14\% & 0.00\% & 0.06\% \\
 &  &  & Abl.2: no adaptive exploration & 0.02\% & 0.12\% & -0.00\% & 0.05\% \\

\end{longtable}

\subsection{Ablation Analysis}\label{subsec:ablation}

Table~\ref{tab:ablation_results} evaluates two controlled ablations of TT-DAC-PS. The first ablation removes the Twin-Target mechanism by replacing the two-target pessimistic backup with a single target critic. The second ablation removes the adaptive exploration design and related policy-smoothing components. Both variants are evaluated under the same execution constraints and deterministic test protocol as the full model.

Ablation~1 shows that the Twin-Target mechanism is an important stabilising component of TT-DAC-PS. Removing the second target critic and the pessimistic minimum backup worsens performance on most instruments. The degradation is especially visible in the more challenging instruments: for AVGO, mean IS increases from \(0.12\%\) for the full model to \(0.47\%\), while for ADBE it increases from \(0.01\%\) to \(0.24\%\). Similar deteriorations are observed for INTC, AMD, PYPL, PEP, and CSCO. These results support the role of the Twin-Target mechanism in reducing unstable value estimates during repeated execution decisions.

Ablation~2 shows that the adaptive exploration design also contributes to robust performance. Removing OU-based adaptive exploration, the deterministic and variance-aware noise adjustment, the SAC-style temperature mapping, and policy smoothing generally weakens execution quality relative to the full model. The largest effects again appear in the more challenging instruments: AVGO increases from \(0.12\%\) to \(0.29\%\), and ADBE from \(0.01\%\) to \(0.14\%\). Smaller deteriorations are observed for INTC, PYPL, PEP, and BKR. This suggests that adaptive exploration and smoothing are particularly useful when volatility, impact uncertainty, or liquidity variation make the execution problem more difficult.

The two ablations therefore play different roles. Ablation~1 primarily tests the contribution of the Twin-Target pessimistic backup to value-estimation stability. Ablation~2 tests the contribution of adaptive exploration and policy smoothing to policy robustness. The stronger deterioration under Ablation~1 for AVGO and ADBE suggests that stable target-value estimation is especially important in high-variance execution settings, while the results for Ablation~2 indicate that adaptive exploration helps the model avoid brittle policies when market conditions vary across episodes.

The effects are not uniform across all instruments. For CMCSA and ORCL, both ablated variants remain close to the full model, suggesting that simpler stabilisation may be sufficient in relatively benign liquidity regimes. By contrast, for AVGO and ADBE, removing either component leads to a substantial increase in implementation shortfall. This heterogeneity supports the view that the full TT-DAC-PS design is most valuable under more difficult microstructure conditions.

Overall, the ablation study supports the proposed architecture. The full TT-DAC-PS model achieves the lowest or near-lowest mean IS across most instruments, while both simplified variants tend to degrade execution quality. These results indicate that the observed performance gains arise from the combination of Twin-Target value stabilisation and adaptive exploration, rather than from a single isolated modelling choice.

\section{Conclusions and Future Work}

\subsection{Conclusions}

This study presented TT-DAC-PS, a deterministic actor-critic framework enhanced with Twin-Target estimation and Policy Smoothing, designed specifically for the optimal trade execution problem. By integrating a hybrid adaptive Ornstein-Uhlenbeck noise schedule, the model addresses the critical challenge of exploration-exploitation balance in non-stationary financial environments. The experimental results across ten U.S. equities demonstrate that the proposed architecture consistently achieves superior execution quality compared to classical baselines and standard on-policy reinforcement learning algorithms.

The key finding is that the performance of TT-DAC-PS is driven by the interaction between Twin-Target value stabilisation and adaptive exploration. The ablation study shows that removing the second target critic and pessimistic minimum backup worsens execution performance on several instruments, particularly AVGO and ADBE, indicating that the Twin-Target mechanism contributes to stable value estimation. Removing the adaptive exploration and policy-smoothing components also weakens results, especially in more difficult volatility and liquidity regimes. Furthermore, the rigorous benchmarking confirms that while classical strategies such as VWAP remain viable in highly liquid, low-volatility regimes, they fail to protect against implementation shortfall in complex, high-impact scenarios where adaptive control is paramount.

\subsection{Future Work}
Several avenues for future research emerge from this work. First, the current market impact model, while grounded in the Almgren-Chriss framework, assumes linear permanent impact. Future iterations will incorporate non-linear transient impact kernels and Hawkes process-driven order flow simulations to better capture the decay of market impact over time. Second, the single-asset execution problem can be extended to a multi-asset portfolio liquidation context, where cross-impact and asset correlations must be managed simultaneously. Finally, future research will investigate risk-aware objective functions beyond the mean-variance framework, such as optimising for the Conditional Value-at-Risk (CVaR) of the shortfall distribution, to better align with the risk preferences of institutional investors.

\section*{Acknowledgments}
The authors gratefully acknowledge the computing time granted by the
KISSKI project.

\bibliographystyle{unsrt}  
\bibliography{references}  

\end{document}